# Lightweight Pedestrian Head-Orientation Recognition Network for Safe Pedestrian-Vehicle Interaction

Yuanzhe Li*, *Graduate Student Member*, *IEEE*, Yidi Huang, Xiaotong Chang, Hounian Liu

***Abstract*—Pedestrian head orientation recognition plays an important role in autonomous driving by providing valuable cues for understanding pedestrian attention and anticipating potential crossing behavior. However, reliable recognition in real-world traffic scenes remains challenging because pedestrian head regions are often captured at low resolution. To address this challenge, we propose a lightweight Low-Resolution Head Orientation Convolutional Neural Network (LRHO-CNN) for pedestrian head orientation recognition. We construct a new dataset by extracting pedestrian head images from multiple public datasets and manually annotating them into eight orientation categories. The collected images are systematically preprocessed and augmented to increase data diversity and better represent variations in illumination and image quality. The experimental analysis compares LRHO-CNN with three fine-tuned baseline models, namely ResNet-18, ResNet-34, and VGG-16. The results demonstrate that LRHO-CNN achieves the highest classification accuracy among the evaluated models. LRHO-CNN is further evaluated on the JAAD and PIE datasets, demonstrating its effectiveness in recognizing pedestrian head orientation in real-world traffic scenes and providing informative head-orientation cues that can support downstream pedestrian behavior and intention prediction.**

***Keywords—head orientation recognition, deep learning, convolutional neural network, pedestrian-vehicle interaction.***

## I. INTRODUCTION

Autonomous vehicles (AVs) have the potential to improve the safety and efficiency of transportation efficiency [1]. However, the dynamic and stochastic nature of pedestrian behavior makes safe pedestrian-vehicle interaction a major challenge for AVs [5, 7, 19]. According to the World Health Organization's Global Status Report on Road Safety 2023 [4], vulnerable road users account for 53% of global traffic fatalities, with pedestrians alone representing 23% of global traffic fatalities. Between 2010 and 2021, global pedestrian fatalities increased by 3% to 274,000 annually, with most fatal incidents occurring during road-crossing events. These findings underscore the importance of reliably interpreting pedestrian behavior, particularly during road-crossing interactions. Early and accurate prediction of pedestrian crossing intention enables AVs to respond in a timely manner, thereby enhancing the safety of pedestrian-vehicle interactions.

Among the behavioral indicators relevant to pedestrian intention prediction, head orientation serves as an early and informative cue of crossing behavior, as pedestrians often turn their heads to monitor approaching traffic before entering the roadway. Pedestrian head-orientation recognition has therefore emerged as an important subtask within the broader field of pedestrian intention prediction for AVs. However, reliable recognition remains challenging in real-world traffic scenes because pedestrian head regions are typically small, low-resolution, and susceptible to occlusion.

To address pedestrian head orientation recognition from small, low-resolution targets, we develop the Low-Resolution Head Orientation CNN (LRHO-CNN), a lightweight architecture designed for low-resolution head images. To enable fine-grained eight-class orientation recognition, we construct a new dataset by integrating samples from the TUD [6], HIIT [18], HOCoffee, and QMUL-Pose-Heads [16] datasets under a unified labeling scheme. Gaussian blur and brightness adjustment are further applied to increase data diversity and simulate image degradation and illumination variations commonly encountered in vehicle-mounted camera images. The main contributions are summarized as follows:

(1) A lightweight convolutional neural network, LRHO-CNN, is developed for eight-class pedestrian head orientation recognition from low-resolution head images.

(2) A new eight-class pedestrian head orientation dataset is constructed by integrating samples from multiple public datasets and manually reannotating them under a unified orientation scheme, with data augmentation applied to simulate variations in image quality and illumination encountered in real-world traffic scenes.

(3) Comparative experiments with fine-tuned ResNet-18, ResNet-34, and VGG-16 models show that LRHO-CNN achieves the highest accuracy. The application of LRHO-CNN to the JAAD and PIE datasets demonstrates its applicability to real-world traffic scenes and its potential to provide informative head-orientation cues.

## II. RELATED WORKS

Pedestrian head orientation provides an informative cue for estimating pedestrians' visual attention and anticipating potential crossing behavior, thereby supporting safer vehicle-pedestrian interaction. It is therefore widely used as an important cue for pedestrian intention prediction, particularly in road-crossing scenarios. Beyond intention prediction, head orientation has also been leveraged in pedestrian tracking and motion modeling. For example, Baxter et al. introduced head-pose estimates as intentional priors in an adaptive Kalman filter, exploiting the correlation between viewing

Chair of Automotive Engineering, Technische Universität Berlin, Berlin, 13355, Germany.

*Yuanzhe Li is the corresponding author of this paper. The E-mail is

direction and pedestrian motion to improve tracking during sudden trajectory changes and occlusions [2].

A growing body of research has investigated head-orientation estimation under different image conditions and representation schemes. Lee et al. proposed a convolutional random projection forest (CRPforest) for head and body orientation estimation from low-resolution images [3]. Their method employs a convolutional random projection network at each tree node to learn discriminative multi-scale filters and compress their responses through sparse random projection, achieving efficient and robust estimation under noise, occlusion, and motion blur. Biternion Nets [11] introduced a continuous angular formulation for head orientation, thereby mitigating the discontinuities inherent in discrete orientation labels. Earlier monocular methods based on handcrafted features [12] achieved promising results in controlled settings but remained sensitive to occlusion and image degradation. Subsequent CNN-based approaches [13-14] improved feature representation and robustness in constrained environments. More recently, Transformer-based methods [15] have been explored to capture long-range dependencies and contextual relationships relevant to pedestrian orientation estimation.

Several public datasets have supported research on pedestrian head orientation recognition, including TUD [6], QMUL-Pose-Heads [16], HOCoffee, and HIIT [18]. However, these datasets differ in image resolution, scene type, camera viewpoint, and annotation protocol. These datasets generally provide relatively coarse orientation labels or are collected under controlled and close-range conditions, making them less representative of the small, low-resolution pedestrian head regions commonly observed by vehicle-mounted cameras. Moreover, limited evaluation has been conducted on real-world pedestrian behavior datasets such as JAAD [9] and PIE [8]. These limitations motivate the construction of a unified eight-class pedestrian head-orientation dataset and the development of a lightweight model tailored to low-resolution pedestrian head-orientation recognition.

## III. Methodology

### A. Problem Formulation

Pedestrian head orientation recognition is formulated as an eight-class classification problem. Given an input head image $X$, the objective is to learn a mapping function $f_\theta$, parameterized by a deep neural network, that produces an eight-dimensional logit vector:

$$z = f_\theta(X) \in \mathbb{R}^8, \quad \hat{y} = \arg\max_{k\in\{1,2,\ldots,8\}} z_k \tag{1}$$

where $\theta$ denotes the learnable model parameters, each element of $z$ corresponds to one predefined head-orientation class, and $\hat{y}$ is the predicted orientation class.

The eight orientation classes are defined as front (f), right-front (rf), right (r), right-back (rb), back (b), left-back (lb), left (l), and left-front (lf). Together, the eight categories divide the full 360° head-orientation range into equal 45° angular intervals.

The inputs consist of low-resolution RGB pedestrian head crops, which are uniformly resized to 50×50 pixels to ensure a consistent input dimension and preserve the low-resolution characteristics commonly observed in vehicle-mounted camera images. The ground-truth label $y$ denotes one of the eight predefined head orientation classes. The network output is converted into a probability distribution using a softmax function, and the orientation class with the highest score is selected as the final prediction. An overview of the pedestrian head orientation classification pipeline is shown in Fig. 1.

### B. Pedestrian Head Orientation Dataset Construction

We construct the pedestrian head orientation dataset using images from four public datasets: HIIT [18], HOCoffee, QMUL [16], and TUD Pedestrian dataset [6]. These datasets span diverse acquisition settings and scene characteristics. HIIT mainly contains low-resolution head images captured against relatively static backgrounds with limited occlusion. QMUL consists of surveillance head images acquired in an airport terminal, introducing variations in subjects, illumination, and background appearance. HOCoffee provides low-resolution outdoor head images collected during coffee-break social interactions, offering more natural interpersonal scenes, whereas the TUD Pedestrian dataset [6] provides pedestrian data captured in real-world street environments. As the TUD Pedestrian samples used in this study do not include head bounding boxes or head orientation labels, a human pose estimation model is employed to localize body keypoints. The detected facial and upper-body keypoin-ts, including the nose, eyes, ears, neck, and shoulders, are then used to estimate the head region and generate the corresponding head crops.

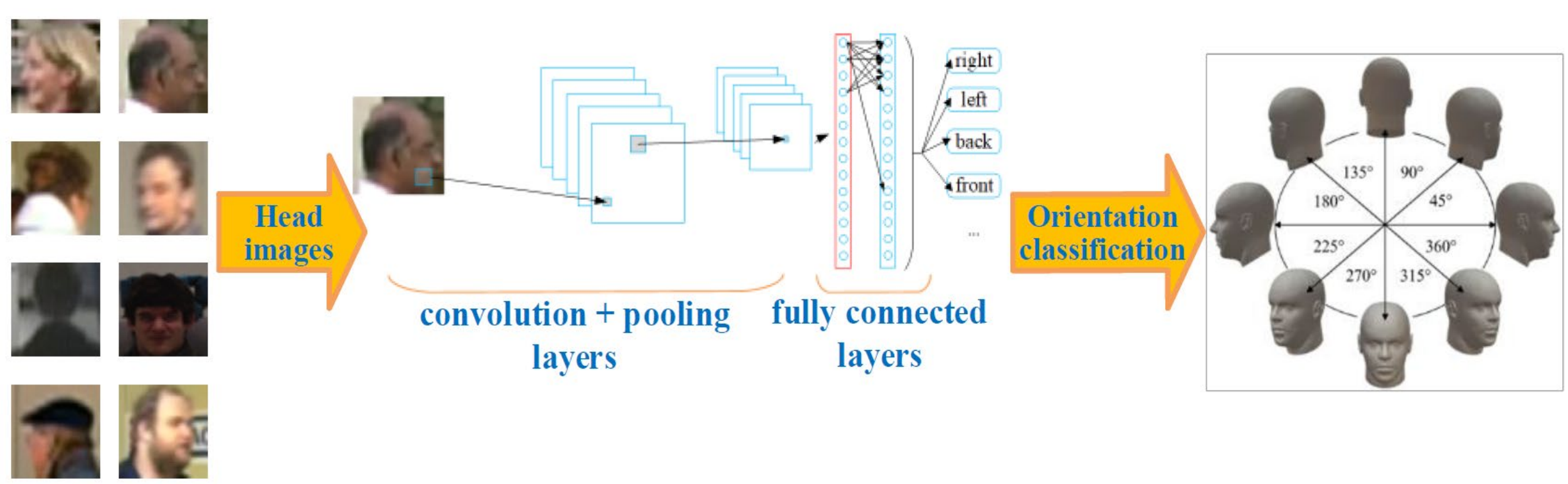

Fig.1. Overview of the pedestrian head orientation classification pipeline.

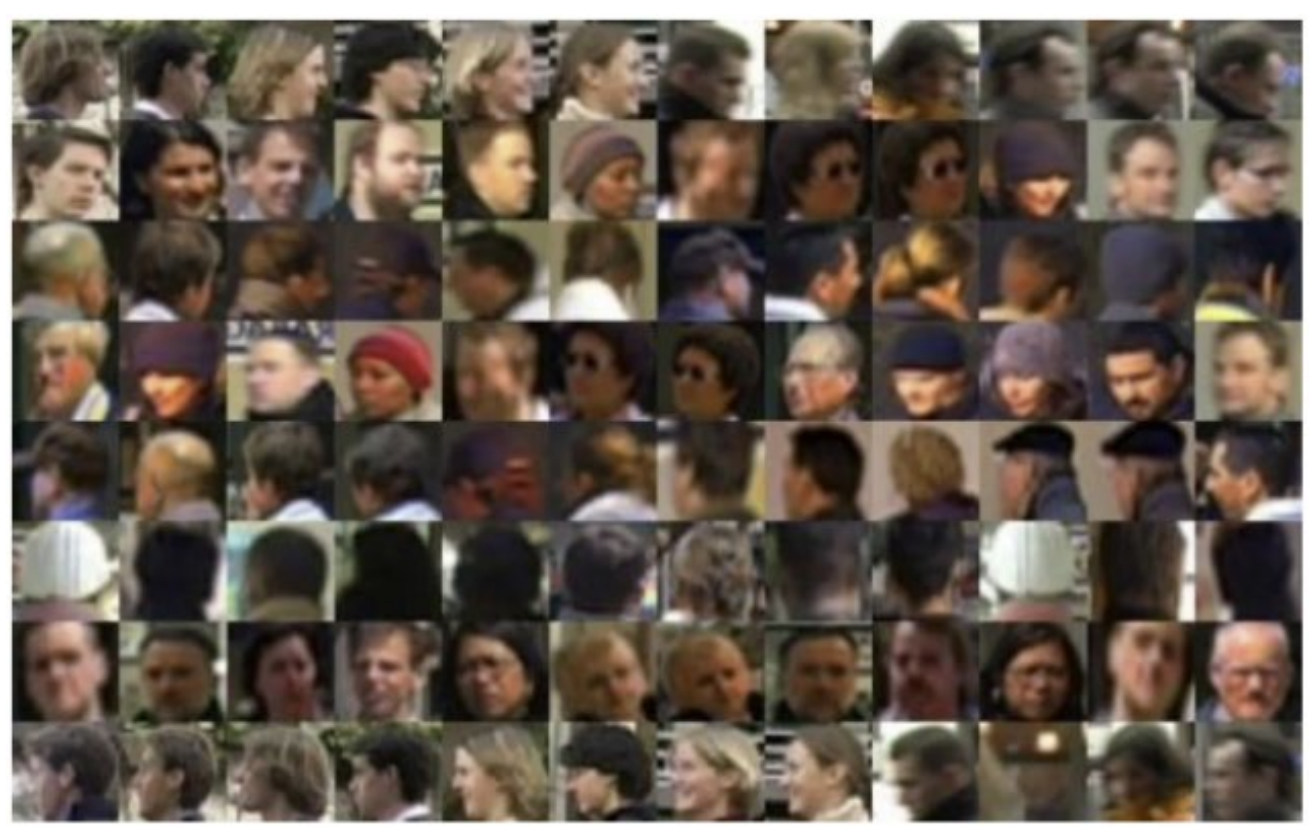

Fig.2. Extracted head images.

All candidate head crops collected from the four source datasets are manually reviewed to remove incorrect detections and samples with insufficient visual quality for reliable orientation annotation. Each retained image is assigned one of eight orientation labels: front (f), right-front (rf), right (r), right-back (rb), back (b), left-back (lb), left (l), or left-front (lf). The annotations are determined primarily from visible head shape and facial orientation, with upper-body posture used as a supplementary cue when necessary. Following the initial annotation, the entire dataset undergoes a second round of manual verification, with particular attention paid to low-resolution, partially occluded, and visually ambiguous samples. The resulting dataset spans a wide range of head postures, illumination conditions, and occlusion sources, including headwear, sunglasses, and partial rear views, reflecting the visual variability encountered in real vehicle-mounted camera footage. Representative sample images drawn from the dataset are shown in Fig. 2.

### *C. Data Preprocessing and Augmentation*

All head images are resized and normalized, followed by data augmentation using Gaussian blur, horizontal flipping, random rotation, and brightness perturbation. Gaussian blur is applied to high-quality subsets such as HIIT [18] to simulate low-resolution degradation.

Each pixel of the blurred image $I_{\text{blurred}}$ is obtained by convolving the original image $I$ with a Gaussian kernel $G$, defined over a local neighborhood of radius $k$ and parameterized by the standard deviation $\sigma$, which controls the degree of smoothing. In this study, $k$=2, corresponding to a 5×5 kernel and $\sigma$ is set to 0.6:

$$I_{\text{blurred}}(x,y)=\sum_{i=-k}^{k}\sum_{j=-k}^{k} I(x+i,y+j)\cdot G(i,j) \tag{2}$$

$$G(i,j)=\frac{1}{2\pi\sigma^2}\exp\left(-\frac{i^2+j^2}{2\sigma^2}\right) \tag{3}$$

Horizontal flipping is employed to generate mirrored samples and mitigate the imbalance between left- and right-facing head orientations. The corresponding class labels are remapped as left ↔ right, front-left ↔ front-right, and back-left ↔ back-right, while the front and back labels remain unchanged. Random rotation is applied within a range of -10° to 10°, with boundary padding used to preserve image integrity along the edges, simulating small natural variations in head posture during data capture.

Brightness adjustment is applied to simulate variable outdoor lighting conditions such as those encountered in real traffic scenes, where a scaling factor $\alpha$ widens or narrows the overall pixel intensity range and an offset $\beta$ shifts the intensity level up or down:

$$I_{\text{enhanced}}=\alpha\cdot I_{\text{original}}+\beta \tag{4}$$

with $\alpha$ sampled from 0.7 to 1.3 and $\beta$ sampled from -40 to 40.

Finally, all images, regardless of their original source or size, are resized to a fixed resolution of 50×50 pixels and normalized to have zero mean and unit variance:

$$X_{\text{norm}}=\frac{X-\mu}{\sigma} \tag{5}$$

where $\mu$ and $\sigma$ represent the mean and standard deviation of its pixel values, respectively, and $X_{\text{norm}}$ denotes the normalized image.

### *D. Lightweight LRHO-CNN Architecture*

As illustrated in Fig. 3, LRHO-CNN comprises four components: an input layer, convolutional modules, pooling layers, and fully connected layers. The input layer receives normalized RGB images of size 50×50×3. The convolutional modules consist of 3×3 convolutional layers with channel dimensions increasing from 100 to 200, enabling fine-grained feature extraction from small targets with a limited parameter count. Several convolutional layers use a stride of 2 for spatial downsampling, while 2×2 pooling layers are inserted between convolutional blocks to further reduce spatial dimensions. After the final convolutional block, global average pooling converts the feature maps into a compact representation, which is fed into a fully connected layer followed by an eight-neuron output layer corresponding to the eight orientation classes. A softmax function produces the class probability distribution:

$$p_i=\frac{\exp(z_i)}{\sum_{j=1}^{8}\exp(z_j)},\quad i\in\{1,2,...,8\} \tag{6}$$

### *E. Training Objective and Optimization*

Pedestrian head-orientation recognition is formulated as a multi-class classification task and optimized using cross-entropy loss. This objective measures the discrepancy between the predicted class probabilities and the ground-truth orientation label, encouraging the network to assign a higher probability to the target class and lower probabilities to the remaining classes.

The dataset is divided into training, validation, and test sets using a group-stratified splitting strategy. Images from the same pedestrian track are assigned exclusively to one su-bset, thereby preventing track-level temporal leakage while approximately preserving the class distribution across all three subsets. The class-wise sample distributions across the eight head-orientation categories in the training and test sets are shown in Fig. 4.

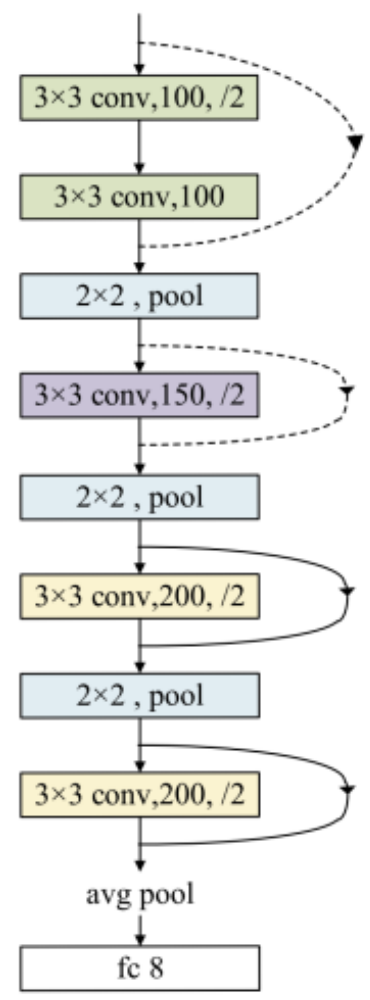

Fig.3. The architecture of LRHO-CNN.

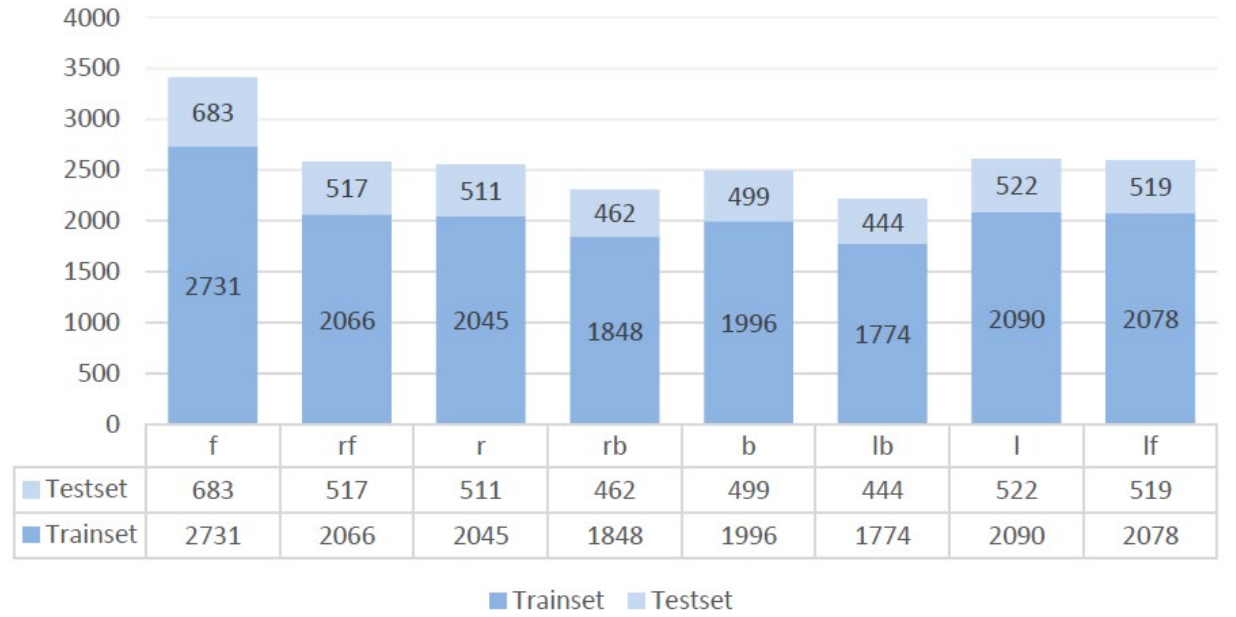

Fig.4. Class-wise sample distributions of the training and test sets across the eight head-orientation categories.

## IV. EXPERIMENTS

### *A. Implementation details*

For a fair comparison, all evaluated models use the same input preprocessing and normalization settings described in Section III-C. The dataset is divided into training, validation, and test sets at an approximate ratio of 8:1:2 using group-stratified splitting strategy. LRHO-CNN is trained from scratch without relying on pretrained weights. ResNet-18 [17], ResNet-34 [17], and VGG-16 [10] are initialized with their standard pretrained weights and adapted to the reduced input resolution and the eight-class classification task. For both ResNet variants, the initial convolutional layer is replaced with a 3×3 convolution with stride 1 to reduce excessive early downsampling of the 50×50-pixel inputs, while retaining 64 output channels. The original classification layers of ResNet-18, ResNet-34, and VGG-16 are replaced with task-specific fully connected layers that output eight-dimensional logit vectors. During training, all layers of the three baseline models are unfrozen and jointly optimized under the same full fine-tuning strategy.

LRHO-CNN is trained using the AdamW optimizer with an initial learning rate of 0.001. Cross-entropy loss is adopted to measure the discrepancy between the predicted class probabilities and the ground-truth labels, imposing larger penalties on confident incorrect predictions:

$$L_{CE} = -\frac{1}{N}\sum_{n=1}^{N}\sum_{c=1}^{8} y_{n,c}\log(p_{n,c}) \tag{7}$$

where $N$ is the batch size, $y_{n,c}$ and $p_{n,c}$ denote the ground-truth label and predicted probability, respectively, for sample $n$ and class $c$.

### *B. Evaluation Metrics*

The classification performance is evaluated using overall accuracy (Acc), macro-averaged precision (Macro-P), macro-averaged recall (Macro-R), macro-averaged F1-score (Macro-F1), and weighted F1-score (Weighted-F1). Acc measures the overall proportion of correctly classified samples. The macro-averaged metrics assign equal importance to all eight orientation classes, thereby reflecting class-balanced performance, whereas Weighted-F1 accounts for differences in class frequency. In addition, confusion matrices are employed to analyze class-wise recognition performance and misclassification patterns among visually similar head orientations.

### *C. Comparative Results*

We first examine the convergence and training stability of LRHO-CNN. As shown in Fig. 5, both the training and valid-

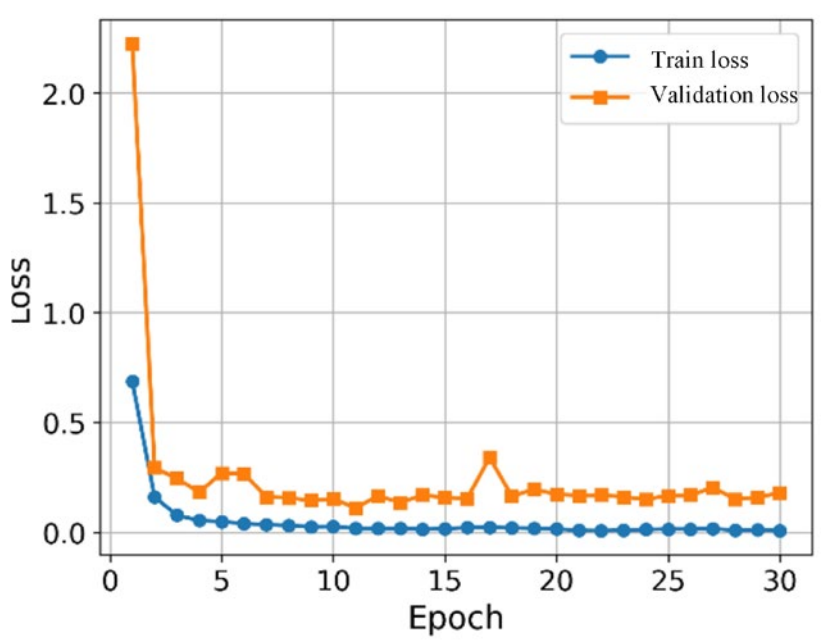

(a) Training and validation loss curves.

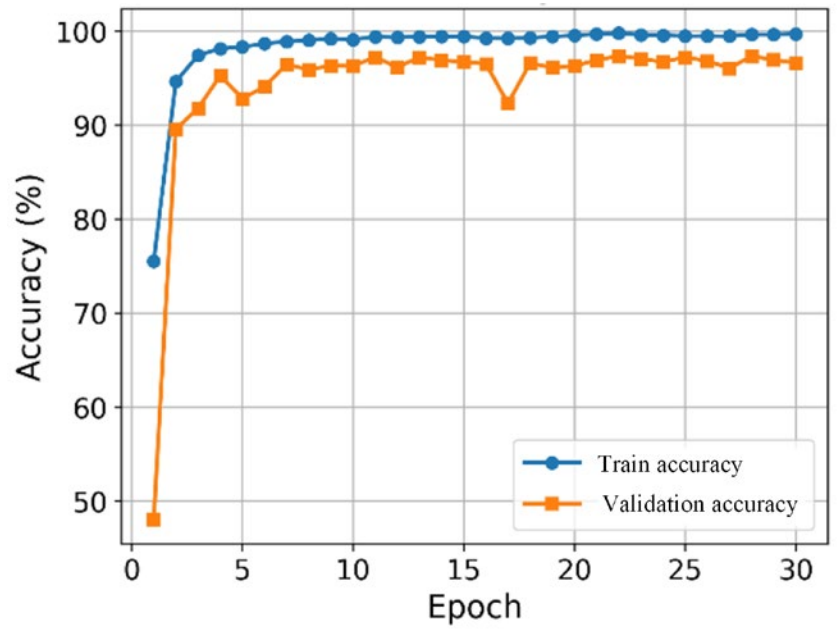

(b) Training and validation accuracy curves.

Fig.5. Training and validation loss and accuracy curves of LRHO-CNN.

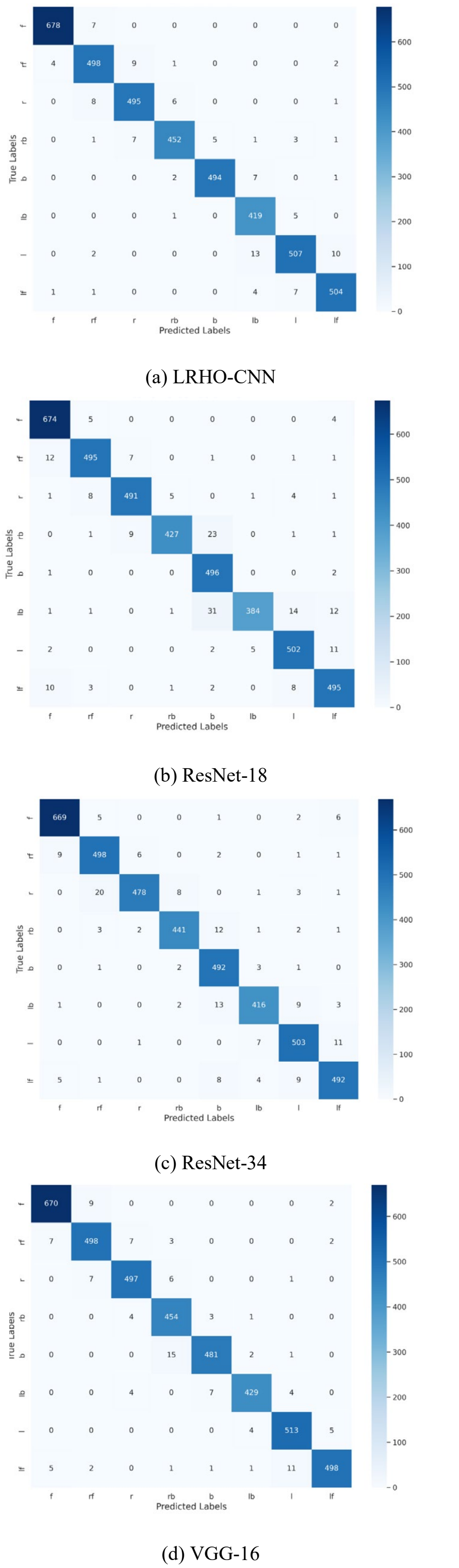

Fig.6. Comparison of confusion matrices for LRHO-CNN and the baseline models.

ation metrics improve substantially within the first five epochs, as evidenced by the pronounced decrease in loss and the corresponding increase in accuracy. The training loss subsequently approaches zero, while the training accuracy stabilizes at nearly 100%. Meanwhile, the validation accuracy remains approximately 96% - 97%, and the validation loss stays at a relatively low level despite minor fluctuations. These results indicate stable optimization and satisfactory generalization performance.

Table I compares LRHO-CNN with ResNet-18, ResNet-34 [17], and VGG-16 [10]. LRHO-CNN achieves the best performance across all metrics, with an accuracy of 97.35%, Macro-P of 97.24%, Macro-R of 97.31%, Macro-F1 of 97.27%, and Weighted-F1 of 97.36%. Compared with the strongest baseline, VGG-16, these metrics improve by 0.12, 0.05, 0.12, 0.09, and 0.13 percentage points (pp), respectively. Although the gains are modest, the consistently higher macro-averaged and weighted scores indicate balanced performance across the eight classes. LRHO-CNN comprises only five convolutional layers and one fully connected layer, resulting in a substantially shallower architecture than the baseline models. Its compact design is well suited to 50×50-pixel head crops and retains sufficient capacity to extract discriminative orientation features. Overall, LRHO-CNN achieves an effective balance between architectural compactness and classification performance.

Fig. 6 compares the confusion matrices of LRHO-CNN and the three baseline models. Although all four models exhibit strong diagonal dominance, most misclassifications occur between visually similar adjacent orientations, such as rf - r, r - rb, rb - b, b - lb, lb - l, and l - lf. ResNet-18 shows notable confusion between lb and b, ResNet-34 between r and rf, and VGG-16 between b and rb as well as lf and l. These errors reflect the difficulty of distinguishing subtle angular differences from low-resolution head images. In comparison, LRHO-CNN produces a concentrated diagonal distribution and fewer errors between adjacent classes, indicating better discrimination of fine-grained head orientation classes.

### *D. Qualitative Results*

To qualitatively assess the applicability and generalization capability of LRHO-CNN in real-world traffic scenes, the trained model is further applied to the JAAD [9] and PIE [8] datasets. Fig. 7 presents representative pedestrian head images from the JAAD and PIE datasets together with their predicted head-orientation labels. Seven examples are provided for each of the eight orientation classes, covering front, back, left, right, and the four diagonal directions. The samples exhibit variations in illumination, apparent head scale, image clarity, background complexity, and partial occlusion caused by headwear or eyewear. Despite these challenging conditions, LRHO-CNN correctly distinguishes clearly defined frontal and rear views as well as more ambiguous transitional orientations, including left-back, right-back, left-front, and right-front. These results qualitatively demonstrate the model's ability to capture discriminative visual cues from small and low-resolution pedestrian head regions in real-world traffic scenes. The resulting eight-class orientation scheme provides a finer-grained description of pedestrian head orien-tation and can provide informative attention-related cues for downstream pedestrian behavior and intention prediction in autonomous driving systems.

TABLE I. COMPARISON OF CLASSIFICATION PERFORMANCE AMONG LRHO-CNN AND BASELINE MODELS

| Method | Acc(%) | Macro-P (%) | Macro-R (%) | Macro-F1 (%) | Weighted-F1 (%) |
|---|---|---|---|---|---|
| ResNet-18 | 95.36 | 95.54 | 95.05 | 95.20 | 95.34 |
| ResNet-34 | 95.96 | 95.94 | 95.84 | 95.87 | 95.96 |
| VGG-16 | 97.23 | 97.19 | 97.19 | 97.18 | 97.23 |
| LRHO-CNN | **97.35** | **97.24** | **97.31** | **97.27** | **97.36** |

Fig.7. Representative qualitative results of LRHO-CNN.

## V. CONCLUSION

In this paper, we presented LRHO-CNN, a lightweight convolutional neural network for eight-class pedestrian head-orientation recognition from low-resolution RGB head crops. The network employs compact convolutional blocks with progressive feature extraction and spatial downsampling to capture discriminative head-orientation cues while maintaining low model complexity. A unified dataset was constructed by integrating and manually re-annotating samples under a consistent eight-class orientation scheme. With data augmentation and a common evaluation protocol, LRHO-CNN outperformed ResNet-18, ResNet-34, and VGG-16, achieving the highest accuracy of 97.35% and reliable recognition of visually similar adjacent orientations. Its application to JAAD and PIE datasets further demonstrates its potential for real-world pedestrian behavior and intention prediction. Future work will investigate temporal head-turning dynamics, more challenging visual conditions, broader cross-dataset evaluation, and computational efficiency.